\documentclass[conference, 10pt]{IEEEtran}
\IEEEoverridecommandlockouts 
\usepackage[]{graphicx}
\graphicspath{{Figures/}}
\usepackage{caption}
\usepackage{subfig} 
\usepackage{amsmath}
\usepackage{amssymb}
\usepackage{epsfig}
\usepackage{cite}
\usepackage{color}
\usepackage{balance}
\usepackage{amsmath}
\usepackage{amsfonts} 
\usepackage{graphicx}
\usepackage{pdfpages}
\usepackage[T1]{fontenc} 
\usepackage{array}
\usepackage{url}

\usepackage{color}
\usepackage{wrapfig}
\usepackage{lipsum}
\usepackage[hidelinks]{hyperref}
\usepackage{multirow}
\usepackage{tabularx}
\usepackage{tikz}
\usepackage{nth} 
\usepackage{algpseudocode} 
\usepackage{algorithm,algpseudocode}
\usepackage{breqn}
\usepackage{todonotes}
\usepackage{amssymb}
\usepackage{verbatim}
\usepackage{longtable}
\usepackage{multirow}
 
\begin{document}

\title{Stereo Matching Based on Visual \\Sensitive Information}

\author{\IEEEauthorblockN{
Hewei~Wang\IEEEauthorrefmark{1}\IEEEauthorrefmark{2},
Muhammad~Salman~Pathan\IEEEauthorrefmark{2}\IEEEauthorrefmark{3}, and
Soumyabrata Dev\IEEEauthorrefmark{2}\IEEEauthorrefmark{3}\IEEEauthorrefmark{4}
}
\IEEEauthorblockA{\IEEEauthorrefmark{1} Beijing University of Technology, Beijing, China}
\IEEEauthorblockA{\IEEEauthorrefmark{2} School of Computer Science, University College Dublin, Ireland}
\IEEEauthorblockA{\IEEEauthorrefmark{3} ADAPT SFI Research Centre, Dublin, Ireland}
\IEEEauthorblockA{\IEEEauthorrefmark{4} Beijing-Dublin International College, Beijing, China}

\thanks{This research has received funding from the European Union's Horizon 2020 research and innovation programme under the Marie Skłodowska-Curie grant agreement No. 801522, by Science Foundation Ireland and co-funded by the European Regional Development Fund through the ADAPT Centre for Digital Content Technology grant number 13/RC/2106_P2.
}
\thanks{Send correspondence to S.\ Dev, E-mail: soumyabrata.dev@ucd.ie.}
\vspace{-0.6cm}
}

\maketitle

\begin{abstract}
The area of computer vision is one of the most discussed topics amongst many scholars, and stereo matching is its most important sub fields. After the parallax map is transformed into a depth map, it can be applied to many intelligent fields. In this paper, a stereo matching algorithm based on visual sensitive information is proposed by using standard images from Middlebury dataset. Aiming at the limitation of traditional stereo matching algorithms regarding the cost window, a cost aggregation algorithm based on the dynamic window is proposed, and the disparity image is optimized by using left and right consistency detection to further reduce the error matching rate. The experimental results show that the proposed algorithm can effectively enhance the stereo matching effect of the image providing significant improvement in accuracy as compared with the classical census algorithm. The proposed model code, dataset, and experimental results are available at
\url{https://github.com/WangHewei16/Stereo-Matching}. 
\end{abstract}

\section{Introduction}

\label{introduction}
With the rapid development of Artificial Intelligence (AI), computer vision as an interdisciplinary subject has widely been used in advertising~\cite{hossari2018adnet,dev2019localizing}, product placement~\cite{nautiyal2018advert,bacher2020advert}, remote sensing~\cite{dev2016estimation,dev2016detecting}, atmospheric study~\cite{manandhar2017analyzing,dev2019multi}, quality control~\cite{birla2015efficient,kaloni2021impact}, biometrics~\cite{patil2020dynamic}, and amongst others. Eyes are the main source of capturing external information and information conveyed by vision is most abundant and intuitive. The visual information reflected by objects is transmitted to the brain through nerve tissue and then processed by the visual cortex of the brain with respect to the content and space respectively. The human eye has a strong ability to distinguish color information and has a great subjective influence, but the perception of gray level image is very poor. The emergence of machine vision can tackle this problem where the resolution of gray reaches 256 levels and has low environmental requirements. Stereo vision technology a sub field of computer vision has always been one of the hot spots of research. During the recent years, stereo matching technology was one of the discussed topics in research community. The depth map obtained by stereo matching technology is widely used in many intelligent fields. The depth information contained in each pixel of the depth map reflects the three-dimensional information of the scene, which has important application value for robot map construction (three-dimensional reconstruction), obstacle avoidance navigation and many other fields. The image source of this technology is usually captured by a binocular camera \cite{Jin2019Enhancing}, which can provide the left and right perspectives of the image, to carry out the stereo matching operations.

\subsection{Related Works}
With the recent developments in stereo matching technology, many research methodologies have been published in this area. Ysimoncelli \textit{et al.} proposed a very classic Sum of Absolute Differences (SAD) stereo matching algorithm based on the difference between two gray regions \cite{Simoncelli1991Probability}. Using the similar method, Da \textit{et al.} furthermore improved the algorithm and used the gray features of the image to find the optimal matching region \cite{Da2013Stereo}. Some scholars use the whole stereo matching image to meet the requirements of the parallax map. In Jian Sun's work \cite{Stereomatching}, a single direction stereo matching algorithm is proposed, which limits the constraints of matching between two images and finds the optimal pixels. Yin Chuanli \textit{et al.} segmented the image by color threshold, fitted the segmented region, and optimized the parallax results by minimizing the energy function. However, the threshold used for color segmentation was very strict, which can easily cause over-segmentation and under segmentation, \cite{improved}. Similar to SAD and Normalized Cross Correlation (NCC) feature extraction methods, Wang \textit{et al.} constructed a real-time matching system, which can get a disparity image quickly, but the matching effect was poor, which can affect the subsequent application \cite{Fei2016The}. Furthermore, Savoy \textit{et al.} in \cite{savoy2017stereoscopic,savoy2016geo} used ground-based sky cameras in a stereo setup to compute the cloud-base height.

\subsection{Contributions of the paper}
According to the previous research done \cite{Non-parametric} on the stereo matching field, the algorithms based on the feature points often can not get depth of the image, and the classical SAD algorithm also needs to be improved for the matching effect. Keeping this point in mind, this paper proposes an image stereo matching algorithm based on visual sensitive information. Ensuring the matching effects, it can further refine the visual sensitive information such as edge points to reduce the false matching rate. 

The main contributions of this paper are as follows:
\begin{itemize}
	\item We provide a systematic analysis of the related frameworks on stereo matching and identify the key shortcomings in them;
	
	\item A novel stereo matching algorithm is proposed wherein we introduce the visual sensitivity factor in the cost calculation stage of the matching process to improve the matching degree; and
	
	\item All the codes related to the different experiments in the paper are open-source\footnote{In the spirit of reproducible research, all the source code is available at \url{https://github.com/WangHewei16/Stereo-Matching}.}. 
\end{itemize}

Primarily, the visual sensitivity factor is introduced in the cost calculation stage of the matching process to improve the matching degree of the contour area. During the cost aggregation stage, a multi-angle extension template based on visual factors is constructed to guide the whole aggregation process. Finally, through the experiments performed on the Middlebury image dataset, the effectiveness and superiority of this method is proved by both the subjective and objective means.

\section{Proposed method of stereo matching}

\begin{figure}[h] 
\centering
\includegraphics[width=0.45\textwidth]{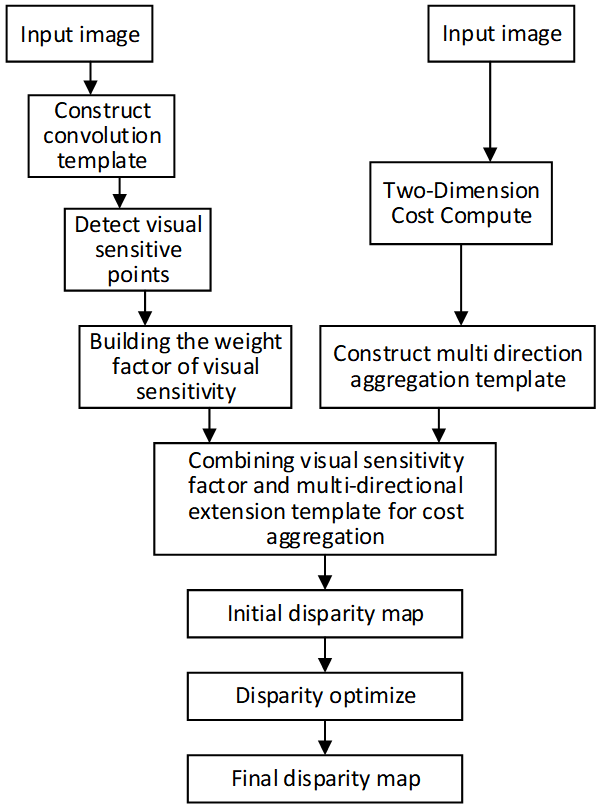} 
\caption{The proposed algorithm for stereo matching. We perform two operations on the image: (a) template convolution operation to get the visual sensitivity factor;  (b) calculation of the initial cost of the image and construct the multi-directional aggregation template. After combining the visual sensitivity factor with the aggregation template, the initial disparity map is generated by the cost aggregation function. Finally, the disparity map is optimized to get the optimized disparity map.}
\label{fig3:Proposed Alogrith flow chart}
\end{figure}

The main  objective of the proposed algorithm is to detect the edge points, contour points and  other visual  sensitive  areas  in  the  image. Furthermore, it analyzes the texture distribution difference between mentioned areas and smoothes them, and guides the judgment weight of degree of freedom extension in the subsequent aggregation stage. In the aggregation stage, the multi-directional extension  template is  used  to process the  covered  pixels,  and  the  extended  pixel  length  is  then guided according  to the  visual sensitivity to  complete  the  whole image processing. The main flow chart is as shown in Figure \ref{fig3:Proposed Alogrith flow chart}.

\begin{figure}[htb]
    \centering
    \includegraphics[width=0.45\textwidth]{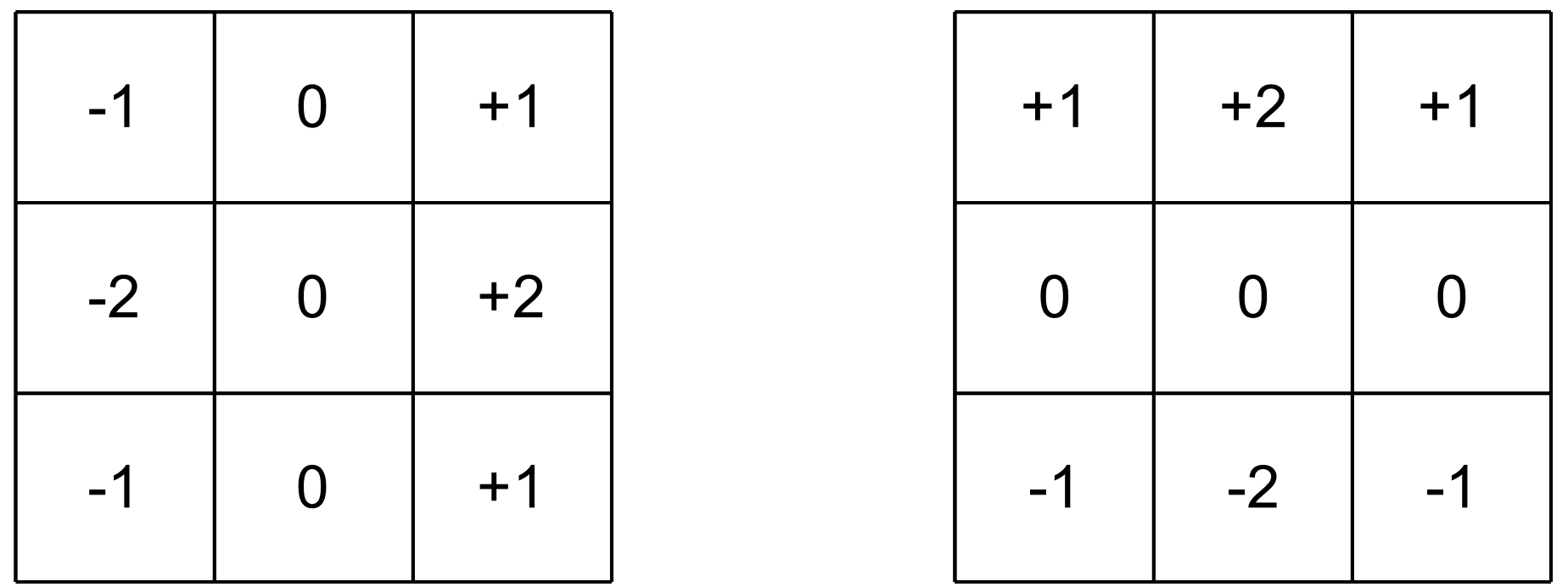} 
    \begin{center}
    \large
    $H_x$\qquad\qquad\qquad\qquad\qquad$H_y$
    \large
    \end{center}
    \caption{Convolution template graph. The template has vertical and horizontal directions. $H_x$ is the horizontal convolution template and $H_y$ is the vertical convolution template. Convolute a region of the image in two directions to get a new factor representing the region.}
    \label{fig4: Convolution template graph}
\end{figure}

After taking the left and right view images as inputs, the step-by-step description of the proposed method is as follows. Firstly, a convolution template is constructed to convolute the covered pixels. The template image is shown in Figure \ref{fig4: Convolution template graph}.

 The template has the vertical and horizontal directions. $H_x$ is the horizontal convolution template whereas the $H_y$ is the vertical convolution template. After convolving with the covered pixels, two important factors can be obtained. One is the brightness difference value of the two directions, and the other is the gradient directivity in the region. Through these two factors, whether the region is a visual sensitive region can be judged. The brightness difference value is calculated as follows:
  
  \begin{equation}
  G_x=H_x*A 
  \end{equation}
  
  \begin{equation}
  G_y=H_y*A 
  \end{equation}
   
   Where, A is the input image, $G_x$ is the transverse brightness value and $G_y$ is the longitudinal brightness value. By summing the absolute values of these two parameters, the gray value of a "point" representing the region can be obtained as follows:
   
   \begin{equation}
   G = \sqrt{G_x^2+G_y^2}
   \end{equation}
   
 Taking G as the second generation value to guide a certain region matching between two images, the first cost is to use the binary string generated according to the gray difference whose hamming distance represents the gray approximation of the two regions. When taking G as the visual sensitivity between two regions as the second condition of image matching, it further judges whether the difference value of G between two regions is less than a certain threshold. Not only the gray approximation should be satisfied, but also the sensitivity must be approximate, then the two regions are considered as candidate matching regions, and then the parallax values of the two regions are calculated. This calculation of the second cost can further constrain the matching conditions so that the accuracy of the matched region is higher.
   
   The process of using two-generation values to judge in turn is represented by the following formula:
   
   \begin{equation}
   f_s(p,d) = 1-exp[-\frac{C_s(p,d)}{\lambda_s}] 
   \end{equation}
   
   \begin{equation}
   \Rightarrow f_G(p,d) = 1-exp\left [ -\frac{C_g(p,d)}{\lambda _g} \right ] 
   \end{equation}
     
   where, p represents the current pixel, d is the horizontal displacement difference of the points in the images, and $C_s$, $C_g$ are the cost of the points in the two positions. $f_s$ is the mapping value of $C_s$ after a certain transformation, also called as gray approximation. After gray screening, the second cost value, $f_g$ is calculated and screened for the visual sensitivity. $\lambda$ is an empirical parameter, which projects the generational values int the interval of 0 and 1.
   
   The initial disparity map of the two images can be calculated by using the method mentioned above, but we can not get the optimal results. In stereo matching technology, the second stage i.e. the cost aggregation, is usually regarded as an important step. In this paper, we propose a cost aggregation method based on a non-directional growth template controlled by visual sensitive directional factor.
   
   According to the calculation method of the second generation value mentioned above, we can easily get the direction of visual sensitivity in a certain area. The calculation is as follows:

   \begin{equation}
   \theta = \arctan (\frac{G_y}{G_x})
   \end{equation}

   This angle is called as the visual sensitivity factor of a target pixel. The visual sensitivity direction of different position points represents the direction of the fastest gray change, and each point has such a direct factor. In this paper, the direction factor is used to guide the cost aggregation between regions in the aggregation phase. A multi-degree freedom growth template guided by direction factor is designed. 
   
   Firstly, during the process of traversing the image, each point is used as a reference point to establish a cross weighted template to cover more regions with similar texture features. The establishment of the weighted template should meet the four degrees of freedom growth decision conditions. This process is expressed as follows:
   
   \begin{equation}
   D_s(p_1, p)<L_1
   \end{equation}
   
   \begin{equation}
   D_c(p_1,p)<f(\tau _1,\theta )
   \end{equation}
   
   \begin{equation}
   D_c(p_1,p_1+(1,0))<\tau _2
   \end{equation}
   
   \begin{equation}
   D_c(p_1,p)<f(\tau _3,\theta ),if\ L_1<D_3(p_1,p)<L_2
   \end{equation}
   
   \begin{equation}
   (\tau _3<\tau _2<=\tau _1,L_1<L_2)
   \end{equation}

   In the formula, P represents the reference point, and $P_1$ represents the position point when it reaches/extends in a certain direction. The first function's $D_s$ is used to calculate the distance between the reference point and the position point. We need to ensure that the range of the extended point is within a reasonable distance.
   
   The second degree of freedom judgment condition $D_c$ represents the judgment condition of the gray level approximation between the extended position point and the reference point. The judgment process is related to the visual sensitivity factor $\theta$. when the extended direction is consistent with or close to the direction of $\theta$, it indicates that the gray level change rate of the direction is large. Therefore, the gray level threshold $\gamma_1$ can be adjusted. Only when the extended position point is still less than the threshold, it is added to the template overlay within the scope. So that the $L_1$ represent the reasonable distance. 
   
   The third decision condition is to compare the color of the new extended position point with the position point at the previous time so that it is less than a certain threshold $\gamma_2$. If this condition is satisfied, the template will continue to be extended and expanded.
   
   The fourth decision condition is that when the extended distance exceeds the reasonable range, we will re-specify the gray threshold $\gamma_3$, which is less than the previous gray threshold. In the same way, when the extension continues, it is still regulated by the direction factor, and a maximum distance parameter will be set at this time. When the extended distance exceeds the threshold, the extension will be terminated immediately.

\section{Experiments and Discussion}

\subsection{Dataset}
Middlebury official test dataset is utilized in this paper for the experimentation. This dataset have also used in other many literature \cite{Kwak2013Implementation}\cite{ZhangFang}. The dataset mainly used for computer vision processing of binocular images including image restoration, stereo matching, image recognition and other. Because of space constraint, this paper mainly selects four kind of images from Middlebury 2.0 for experiments containing 24 left and right view color images and the real parallax images. The proposed algorithm is tested on windows 7, 64 bit operating system. Microsoft Visual Studio and Open CV2.4.9 framework were used for experiments. 

\subsection{Subjective evaluation}

To clearly describe the concept of mismatch rate and the effectiveness of the proposed algorithm, this paper presents a subjective form to show the image matching effect and compares the test image obtained by proposed algorithm with the test image obtained by the classical algorithm.

\begin{figure*}[htb]
\centering
\includegraphics[width=0.9\textwidth]{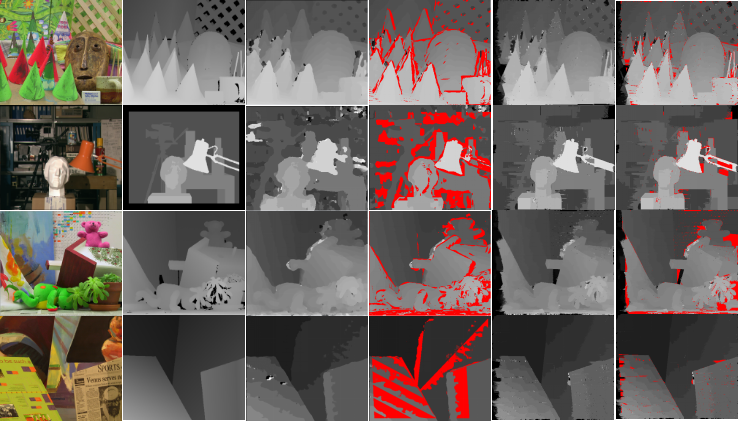} 
\caption{Comparison of different test results. The first column is the perspective image of the original image from the experimental data set, the second column is the ideal parallax image for reference, the third and fourth columns are the resultant images obtained by the classical algorithm census, and the fifth and sixth columns are the images obtained by stereo matching algorithms based on visual sensitive information. Among them, the red marks in the fourth and sixth columns indicate the mismatching areas when comparing the two resultant maps with the ideal disparity map.}
\label{fig7:Comparison of different test results}
\end{figure*}

Figure \ref{fig7:Comparison of different test results} shows the comparison results of the two, mainly for the four types of images from the mentioned data set. From top to bottom, the image categories are cones, Tsukuba, Teddy and Venus. The images are displayed into seven columns. After the experimentation performed by using two algorithms, the comparison results are shown, and a more detailed description is given in the figure below.

\subsection{Objective evaluation}
Next, we have showed the objective data to prove the effectiveness of the proposed algorithm. Table \ref{tab1} shows the error matching rate of the four types of images processed by the proposed algorithm and compares with other matching algorithms of the same kind.

\begin{table}[htb]
\centering
\caption{The error matching rate obtained by various algorithms}
\begin{tabular}{c|ccccc}
\hline
\textbf{Algorithm} & \textbf{cones} & \textbf{tsukuba} & \textbf{teddy} & \textbf{venus} & \textbf{avg\%}\\
\hline
Census & 21.89 & 27.41 & 23.43 & 27.78 & 25.13 \\
Proposed & 4.10  & 2.01  & 7.84  & 1.35  & 3.82 \\ 
Reference\cite{Optics} & 4.04  & 2.53  & 7.57  & 1.60  & 3.93 \\ 
SG-C\cite{Guo} & 12.92 & 4.80  & 8.05  & 1.91  & 6.92 \\ 
Mp-C\cite{ZhangFang} & 6.78  & 4.50  & 11.32 & 3.55  & 6.53 \\ 
\hline 
\end{tabular}
\label{tab1}
\end{table}

These values are obtained by comparing them in the non-occluded area. The non-occluded region in the left and right view of the images is an area which matches in both the views without any event/object disturbance, we call this area a non-occluded region. Occlusion detection refers to the set of techniques employed to detect which areas of the images are occlusion boundaries or areas that appear occluded in views of the scene. Generally speaking, only comparing the matching effect of non-occluded regions is better than that of all image regions. It can be seen from Table \ref{tab1} that the error matching rate of the proposed algorithm is lower than that of the same kind of algorithm, and the overall matching effect is better

\subsection{Impact of noise on mismatch rate}
The algorithm proposed in this paper is more robust than the compared works. To prove this, we added different concentrations of Gaussian noise to the above four original images, and it is mainly expressed by the change of average mismatch rate with the increase of noise concentration. The proposed method also uses the improved adaptive aggregation template.  The comparison results are shown in Table \ref{tab2}, where “none” represents the case without the input noise. The results in the table proves that when the noise concentration increases, the change in the degree of error matching rate and the sensitivity to noise remains low.

\begin{table}[htb]
\small 
\centering
\caption{Mismatch rate under different Gaussian noise concentrations}
\begin{tabular}{c|ccccc}
\hline
\textbf{Noise} & \textbf{none} & \textbf{2\%} & \textbf{5\%} & \textbf{10\%} & \textbf{15\%}\\
\hline
Ref. \cite{Chai2} & 4.54 & 6.67 & 11.73 & 26.88 & 48.63 \\
Proposed & 3.82 & 5.02 & 8.65 & 17.33 & 38.46 \\
\hline 
\end{tabular}
\label{tab2}
\end{table}

\section{Conclusion \& Future Work}
In this paper, the stereo matching technology is studied, and a stereo matching algorithm based on visual sensitive information is proposed. In the cost calculation stage, the sensitive information factor is proposed, and the robustness of cost matching is enhanced. In the process of aggregation, the aggregation template is extended for each pixel according to the sensitivity factor. Finally, in the experimental part, the proposed algorithm is demonstrated subjectively and proved by objective data through the combination of graph and tables.Summing up, the proposed algorithm can effectively match the left and right view of images, and the matching effect is better than a variety of similar algorithms.

For the future work, we intend to further study the factors that have a direct impact on the parallax value, and the actual scene image depth map generation and three-dimensional environment reconstruction work, as far as possible to capture more actual scene images to form their data set.	

\balance


\end{document}